\documentclass[conference]{IEEEtran}
\IEEEoverridecommandlockouts
\usepackage{cite}
\usepackage{amsmath,amssymb,amsfonts}
\usepackage{algorithmic}
\usepackage{graphicx}
\usepackage{caption}
\usepackage{textcomp}

\usepackage{amsmath,amsfonts,bm}

\def\eqref#1{equation~\ref{#1}}

\def\1{\bm{1}}

\DeclareMathAlphabet{\mathsfit}{\encodingdefault}{\sfdefault}{m}{sl}
\SetMathAlphabet{\mathsfit}{bold}{\encodingdefault}{\sfdefault}{bx}{n}

\usepackage[bookmarks=false]{hyperref}
\usepackage{url}
\usepackage[utf8]{inputenc} 
\usepackage[T1]{fontenc}    

\usepackage{booktabs}       
\usepackage{amsfonts}       
\usepackage{nicefrac}       
\usepackage{microtype}      
\usepackage{fancyvrb}
\usepackage{color}
\usepackage{float}
\usepackage{bm}
\usepackage{amsmath}
\usepackage{graphicx}
\usepackage{csquotes}
\usepackage{xargs}  
\usepackage{subcaption}
\usepackage[acronym,nomain]{glossaries}
\usepackage[normalem]{ulem}
\usepackage[pdftex,dvipsnames]{xcolor}
\usepackage[colorinlistoftodos,prependcaption,textsize=tiny]{todonotes}

\newacronym{mlp}{MLP}{Multi Layer Perceptron}
\newacronym{rbf}{RBF}{\textit{Radial Basis Function}}
\newacronym{rks}{RKS}{\textit{Random Kitchen Sinks}}
\newacronym{rff}{RKS}{\textit{Random Fourier Features}}
\newacronym{svd}{SVD}{Singular Value Decomposition }
\newacronym{relu}{ReLU}{Rectified Linear Unit}
 
\newcommandx{\itodo}[2][1=]{\todo[inline, #1]{#2}}
\newcommandx{\hachem}[2][1=]{\todo[linecolor=red,backgroundcolor=red!25,bordercolor=red,#1]{#2}}
\newcommandx{\thier}[2][1=]{\todo[linecolor=blue,backgroundcolor=blue!25,bordercolor=blue,#1]{#2}}
\newcommandx{\steph}[2][1=]{\todo[linecolor=OliveGreen,backgroundcolor=OliveGreen!25,bordercolor=OliveGreen,#1]{#2}}

\newcommand{\x}{\textbf{x}}

\newcommand{\randcomp}{\textbf{Q}}

\newcommand{\phiff}{\phi_{ff}}
\newcommand{\phinys}{\phi_{nys}}
\newcommand{\phirks}{\phi_{rks}}

\newcommand{\dimconv}{p}
\newcommand{\dimrepr}{q}
\newcommand{\Sff}{\textbf{S}}
\newcommand{\Hadamard}{\textbf{H}}
\newcommand{\Gff}{\textbf{G}}
\newcommand{\Bff}{\textbf{B}}
\newcommand{\Piff}{\boldsymbol{\Pi}}
\newcommand{\K}{\boldsymbol{K}}
\newcommand{\Kapprox}{\tilde{\K}}

\newcommand{\subsize}{m}
\newcommand{\trainsize}{n}
\newcommand{\nyssubsample}{\textbf{L}}
\newcommand{\kernelvector}{\textbf{k}_{\x, \nyssubsample}}
\newcommand{\weightnys}{\textbf{W}}

\newcommand{\randcompff}{\textbf{V}}

\def\BibTeX{{\rm B\kern-.05em{\sc i\kern-.025em b}\kern-.08em
    T\kern-.1667em\lower.7ex\hbox{E}\kern-.125emX}}

\begin{document}

\title{Deep Networks with Adaptive Nyström Approximation\\}

\author{\IEEEauthorblockN{1\textsuperscript{st} Giffon Luc*}
\IEEEauthorblockA{Marseille, France \\
luc.giffon@lis-lab.fr}\\
\and
\IEEEauthorblockN{2\textsuperscript{nd} Ayache Stéphane}
\IEEEauthorblockA{Marseille, France \\
stephane.ayache@lis-lab.fr}
\and
\IEEEauthorblockN{4\textsuperscript{th} Kadri Hachem*}
\IEEEauthorblockA{Marseille, France \\
hachem.kadri@lis-lab.fr}\\
\and
\IEEEauthorblockN{3\textsuperscript{rd} Artières Thierry*}
\IEEEauthorblockA{Marseille, France \\
thierry.artieres@lis-lab.fr}

}

\author{\IEEEauthorblockN{Luc Giffon,
Stéphane Ayache, Thierry Artières and
Hachem Kadri}
\IEEEauthorblockA{\textit{Aix Marseille Université, Universit\'e de Toulon, CNRS, LIS, Marseille, France }\\
firstname.name@lis-lab.fr}
}

\maketitle

\begin{abstract}
Recent work has focused on combining kernel methods and deep learning to exploit the best of the two approaches. 
Here, we introduce 
a new architecture of neural networks in which we replace the top dense layers of standard convolutional architectures with an approximation of a kernel function by relying on the Nyström approximation. 
Our approach is easy and highly flexible. It is compatible with any kernel function and it allows exploiting multiple kernels. 
We show that 
our architecture has the same performance than standard architecture on datasets like SVHN and CIFAR100. One benefit of the method lies in its limited number of learnable parameters which makes it particularly suited for small training set sizes, e.g. from 5 to 20 samples per class. 



\end{abstract}

\section{Introduction}
\label{sec:introduction}

Kernel machines and deep learning have mostly been investigated separately. Both have strengths and weaknesses and appear as complementary family of methods with respect to the settings where they are most relevant. Deep learning methods may learn from scratch relevant features from data and may work with huge quantities of data. Yet they actually require large amount of data to fully exploit their potential and may not perform well with limited training datasets. Moreover deep networks are complex and  difficult to design and require lots of computing and memory resources both for training and for inference. Kernel machines are powerful tools for learning nonlinear relations in data and are well suited for problems with limited training sets. Their power comes from their ability to extend linear methods to nonlinear ones with theoretical guarantees. However, they do not scale well to the size of the training datasets and do not learn features from the data. They usually require a prior choice of a relevant kernel amongst the well known ones, or even require defining an appropriate kernel for the data at hand.


Although most research in the field of deep learning seems to have evolved as a \enquote{parallel learning strategy} to the field of kernel methods, there are a number of studies at the interface of the two domains which investigated how some concepts can be transferred from one field to another. Mainly, there are two types of approaches that have been investigated to mix deep learning and kernels. Few works explored the design of deep kernels that would allow working with a hierarchy of representations as the one that has been popularized with deep learning \cite{cho_saul, montavon2011kernel, jose2013local, heinemann2016improper, steinwart2016learning, wilson2016deep}. Other studies focused on various ways to plug kernels into deep networks \cite{mairal2014convolutional, deepfriedconvnets, hazan2015steps, mairal2016end, zhang2017stacked}.
%
%
This paper follows this latter line of research. Specifically, we propose a new kind of architecture which is built by replacing the top dense layers of a convolutional neural network by an adaptive approximation of a kernel function. A similar approach proposed in the literature is \textit{Deep Fried Convnets}~\cite{deepfriedconvnets} which brings together convolutional neural networks and kernels via \textit{Fastfood}~\cite{fastfood}, a kernel approximation technique based on random feature maps. We revisit this concept in the context of Nyström kernel approximation~\cite{nystrom}. One key advantage of our method is its flexibility that enables the use of any kernel function. Indeed, since the Nyström approximation uses an explicit feature map from the data kernel matrix, it is not restricted to any specific family of kernel function. 
 This is also useful when one wants to use or learn multiple different kernels instead of a single kernel function, as we demonstrate here. In particular we investigate two different ways of using multiple kernels, one is a straightforward extension of the single kernel version while the second is a 
variant that exploits a Nyström kernel approximation for each of the feature map output of the convolution


Our experiments on four datasets (MNIST, SVHN, CIFAR10 and CIFAR100) highlight three important features of our method. First our approach compares well to standard approaches in standard settings (using full training sets) while requiring a reduced number of parameters compared to full deep networks. 
This specific feature of our proposal makes it suitable for dealing with limited training set sizes as we show by considering experiments with tens or even fewer training samples per class.
Finally the method may exploit multiple kernels, 
providing a new tool with which to approach the problem of Multiple Kernel Learning~(MKL)~\cite{MKL}, and enabling taking into account the rich information in multiple feature maps of convolution networks through multiple Nyström layers. 


It should be noted that some recent works has focused on using the Nyström approximation in conjonction with neural networks. In \cite{mairal2016end}, Nyström method was used  to enable tractable computation in convolutional kernel networks by projecting data features in a space of low dimension. In \cite{croce2018explaining}, it was used to transform  input data before feeding them to a neural network in order to achieve better interpretability. Finally,  \cite{song2018optimizing} used an ensemble of Nyström approximation and then applied a neural network to optimize a kernel machine. To our knowledge, the Nyström approximation itself has not been studied as a drop-in replacement for standard fully-connected layers in deep  networks neither it has been used to adaptively learn a metric on the feature space induced by the convolution filters, as presented in our work.

The rest of the paper is organized as follows. We provide background on kernel approximation via the Nyström and the random Fourier features methods in~Section~\ref{sec:SOA}. The detailed configuration of the proposed \textit{Adaptive Nyström Newtorks} is described in~Section~\ref{sec:deepstrom}. We also show in Section~\ref{sec:deepstrom} how adaptive Nyström networks can be used with multiple kernels.  Section~\ref{sec:xp} reports experimental results on MNIST, SVHN, CIFAR10 and CIFAR100 datasets to first provide a deeper understanding of the behaviour of our method with respect to the choice of the kernels and the combination of these, and second to compare it to the usual fully-connected layers on classification tasks with respect to accuracy and to complexity issues, in particular in the small training set size setting.

\section{Background on Kernel Approximation}
\label{sec:SOA}

Kernel approximation methods have been proposed to make
kernel methods scalable. Two popular methods are Nyström approximation~\cite{nystrom} and random features approximation~\cite{rff}. The former approximates the kernel matrix by an efficient low-rank decomposition, while the latter is based on mapping input features into a low-dimensional feature space where dot products between features approximate well the kernel function.

\medskip
\paragraph{Nyström approximation~\cite{nystrom}}
It computes a low-rank approximation of the kernel matrix by randomly subsampling a subset of instances. 
%
Let consider a training set of $n$ training samples, $\left\{x_i \in \mathbb{R}^d, i =1,..,n  \right\}$,  $\K$ be the kernel matrix defined as $\K(i,j) = k(\x_i, \x_j), \forall~i,j\in[1, \ldots, \trainsize]$, and $\nyssubsample$ be a subset of examples $\nyssubsample = \left\{\x_i\right\}_{i=1}^\subsize$  which is selected from the training set. Assuming the subset includes the first samples, or rearranging the training samples this way, $\K$ may be rewritten as:
\[
  \K = 
  \begin{bmatrix} 
  \K_{11} & \K_{21}^T \\ 
  \K_{21} & \K_{22} 
  \end{bmatrix},
\]
where $\K_{11}$ is the Gram matrix on subset $\nyssubsample$. 
Nyström approximation is obtained as follows
\[
\K \simeq \Kapprox = 
\begin{bmatrix} 
\K_{11} \\
\K_{21} 
\end{bmatrix}
\K_{11}^{-1}
\begin{bmatrix} 
\K_{11} & \K_{21} ^T
\end{bmatrix}.
\]
From this approximation the Nyström nonlinear representation of a single example $\x$ is given by 
\begin{equation}
\label{eq:nys}
\phinys(\x) =  \kernelvector \K_{11}^{-\frac{1}{2}},
\end{equation}
where $\kernelvector=[k(\x, \x_1), \ldots, k(\x, \x_m)]^T$ with $\x_i \in \nyssubsample$.

\smallskip

One key aspect of the Nystr\"om methods is the sampling technique used to select informative columns from $\K$. It influences the subsequent approximation accuracy and thus the performance of the learning algorithm. See~\cite{kumar2012sampling,sun2015review} for more details and a comparison of various sampling methods for the Nystr\"om approximation. It is of note that uniform sampling was adopted when the standard Nystr\"om method was introduced~\cite{nystrom}. It is a widely applied sampling method due to its low time consumption~\cite{sun2015review}.

\medskip
\paragraph{{Random features approximation}~\cite{rff}}

It computes a low-dimensional feature map  $\tilde \phi$ of dimension $q$ such that $\langle \tilde \phi(\cdot),  \tilde \phi(\cdot) \rangle = \tilde k(\cdot, \cdot) \simeq k(\cdot, \cdot)$. Two well-known instances of this method are \gls{rks}~\cite{rks} and \textit{Fastfood}~\cite{fastfood}. \gls{rks} approximates a \gls{rbf} kernel using a random Fourier feature map defined as
\begin{equation}
\label{eq:rks}
\phirks(\x)=\frac{1}{\sqrt{\dimconv}} [\cos(\randcomp\x) \sin(\randcomp\x)]^T,
\end{equation}

~\\
where $\x \in \mathbb{R}^d$, $\randcomp \in \mathbb{R}^{\dimrepr \times d}$ and $\randcomp_{i,j}$ are drawn randomly. If $\randcomp_{i,j}$ are drawn according to a Gaussian distribution then the method is shown to approximate the Gaussian kernel, i.e.  $\langle \phirks(\x_1), \phirks(\x_2) \rangle \approx \exp(-\frac{||\x_1 - \x_2||^2}{\sigma})$ where $\sigma$ is the hyper-parameter of the kernel. Note that $\sigma$ is related to the parameters of the Gaussian distribution that generates the random features.
Replacing the $\cos$ and $\sin$ activation functions by a \gls{relu} allows to approximate the arc-cosine kernel instead of the \gls{rbf} Gaussain kernel~\cite{cho_saul, pmlr-v32-pandey14}.

The \textit{Fastfood} method \cite{fastfood} is a variant of \gls{rks} with reduced computational cost for the Gaussian kernel. It is based on approximating the matrix $\randcomp$ in Eq.~\ref{eq:rks}, when $q=d$, by a product of diagonal and hadamard matrices according to
%
\begin{equation*}
\label{eq:Vff}
\randcompff = \frac{1}{\sigma\sqrt{d}} \Sff\Hadamard\Gff\Piff\Hadamard\Bff,
\end{equation*}
where $\Sff$,$\Gff$ and $\Bff$ are diagonal matrices of size $d \times d$, $\Piff \in \{0,1\}^{d \times d}$ is a random permutation matrix, $\Hadamard$ is a Hadamard matrix which does not requite to be stored, and $\sigma$ is an  hyperparameter. 
%
%
Matrix $\randcompff$ may be used in place of $\randcomp$ in Eq.~\ref{eq:rks} to define the \textit{Fastfood} nonlinear representation map
\begin{equation}
\label{eq:ff}
\phiff(\x) = \frac{1}{\sqrt{\dimconv}} [\cos(\randcompff\x) \sin(\randcompff\x)]^T.
\end{equation}
Note that this definition requires $d$ to be a power of $2$ to take advantage of the recursive structure of the Hadamard matrix. Note also that to reach a  representation dimension $\dimrepr > d$ one may compute multiple $\randcompff$ and concatenate the  corresponding $\phiff$. 

In the next section we introduce deep adaptive Nystr\"om networks. 
They combine efficiently kernel Nystr\"om approximation with deep learning to learn nonlinear mappings between layers in a neural network architecture while requiring few parameters.






\begin{figure*}
\centering
\includegraphics[width=0.9\textwidth,keepaspectratio]{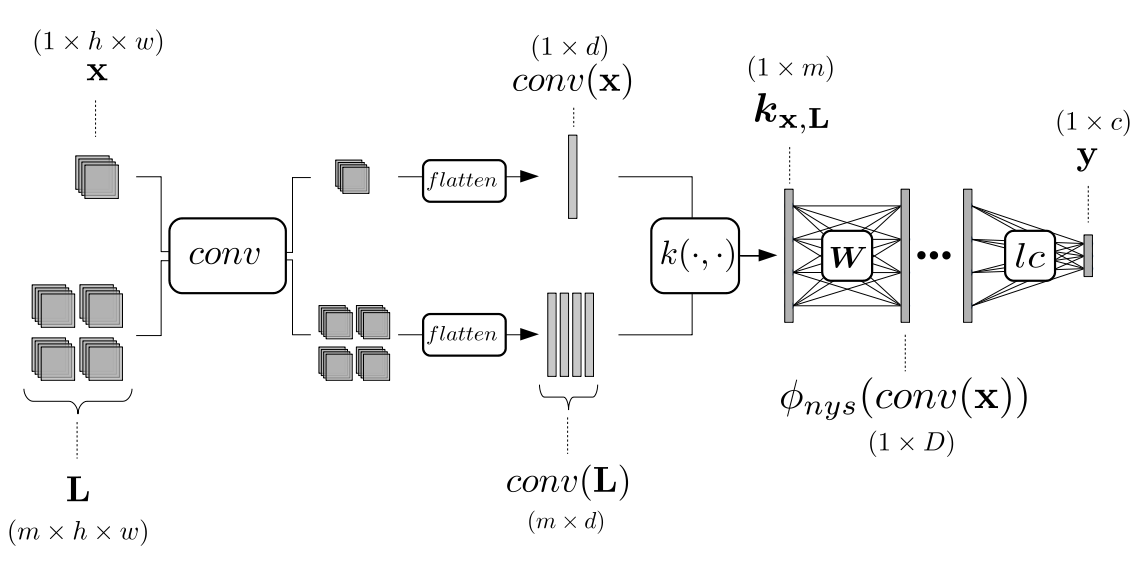}
        \caption{The architecture we propose involves a usual convolutional part, \textit{conv} , including multiple convolutional blocks, and a Nyström layer which is then fed to (eventually multiple) standard dense layers up to the classification layer. The Nyström layer computes the kernel between the output of the $conv$ block for a given input and the corrsponding representations of the trains samples in the subsample $L$ before applying a linear transformation $\weightnys$ to the so obtained kernel vector $\kernelvector$.}
        \label{fig:phinys}
\end{figure*}

\section{Deep Adaptive Nystr\"om Networks}
\label{sec:deepstrom}
Deep adaptive Nyström network is a deep learning architecture that replaces dense layers of a convolutional neural architecture by a \textit{Nytsr\"om} approximation of a kernel function. This allows to take advantage of the low complexity cost in terms of computation and memory of the Nytsr\"om method to reduce significantly the computation cost and the number of parameters of the fully-connected layers in deep convolutional neural networks. 
{It is of note that although we introduce deep Nyström networks in the context of deep convolutional networks, they are general as they can also be applied to other deep neural network architectures, can work with any kernel function or a combination of kernels, and do not require a vectorial representation of the features.}

First, we start by revisiting the concept of Nyström kernel approximation from a feature map perspective.

\paragraph{Nyström approximation from an empirical kernel map perspective}

The empirical kernel map is an explicit $n$-dimensional feature map that is obtained by applying the kernel function on the training input data~\cite{scholkopf1999input}. It is defined as
\begin{align}\phi_{emp} : \quad &\mathbb{R}^d \to \mathbb{R}^n \nonumber \\
& x \mapsto \big( k(x,x_1), \ldots, k(x,x_n) \big).
\end{align}
An interesting feature of the empirical kernel map is that if we consider the inner product in $\mathbb{R}^n$ $\langle \cdot,\cdot\rangle_M = \langle \cdot, M\cdot\rangle$  with  a positive semi-definite~(psd) matrix $M$, we can recover the kernel matrix $K$ using the empirical kernel map by setting $M$ equals to the inverse of the kernel matrix. In other words, $K_{emp} := \big(\langle \phi_{emp}(x_i), \phi_{emp}(x_j) \rangle_{K^{-1}}\big)_{i,j=1}^n = K$.
Since $K$ is a psd matrix, one can consider the feature $\phi_{emp}': x \to K^{-1/2}\phi_{emp}(x)$ as an explicit feature map that allows to reconstruct the kernel matrix. This feature map is of dimension $n$ and then is not interesting when the number of example is large.  
%

From an empirical kernel map point of view, $\phinys(\x)$~(Equation \ref{eq:nys}) can be seen  as an {empirical kernel map}~\cite{scholkopf1999input} and $\K_{11}^{-\frac{1}{2}}$ as a metric in the {empirical feature space}. From this viewpoint, we think that it could be useful to learn a metric $\weightnys$ in the empirical feature space instead of assuming it to be equal to $\K_{11}^{-\frac{1}{2}}$. In a sense, this should allow to learn a kernel by learning its Nyström feature representation. In the following, we call the setting where $\weightnys$ is learned by the network as \textit{deep adaptive Nyström}.

\medskip

\paragraph{Principle}

As illustrated in Figure~\ref{fig:phinys}, the model we propose is based on using the Nyström approximation to integrate any kernel function on top of convolutional layers of a deep net. 

%
One main advantage of our method is its generic feature that enables the use of any kernel, or a combination of them. This is particularly useful if one wants to use multiple kernels or have no prior knowledge on which kernel to use for a particular task.
Starting from 
an usual deep neural network, we replace the top fully-connected hidden layers with $\phinys$ (Equation \ref{eq:nys}). 
In order to compute the above Nyström representation of a sample $\x$ one must consider a subsample $\nyssubsample$ of training instances. Since the kernel $k$ is computed on the representations given by convolutional layers, the samples in  $\nyssubsample$ must be represented in the same space, and hence must be processed by the convolutional layers as well. Once convolutional representations are calculated, the kernel function may be computed with an input sample and each instance in $\nyssubsample$ in order to get the vector $\kernelvector$, which is then linearly transformed by $\weightnys$ before the linear classification layer (see Figure \ref{fig:phinys}). 
However, one problem arises with the computation of  $\textbf{K}_{11}^{-\frac{1}{2}}$ which requires the computation of the \gls{svd} of $\textbf{K}_{11}$ for each batch. In the case where the size of the subsample $\nyssubsample$, $\subsize$,  is relatively large, the computational complexity of the \gls{svd} is  $O(\subsize^3)$ for a single batch whose size is likely to be of the same order of magnitude than $\subsize$. 
This issue can be at least partially settled via  \textit{Adaptative-Nyström} network where, instead of setting the weights  $\weightnys = \K^{-\frac{1}{2}}$ as in Eq. \ref{eq:nys}, we learn these weights as parameters of the network via gradient descent. In this Adaptive-Nyström scheme, the computational and storage complexity of our approach after the convolution are both $O(dm + m^2 + mc)$ against $O(dD + Dc)$ for standard fully-connected layers, with $d$ being the input dimension of the layer, $D$ the output dimension and $c$ the number of classes.

\begin{figure*}
    \centering
\includegraphics[width=0.4\textwidth,keepaspectratio]{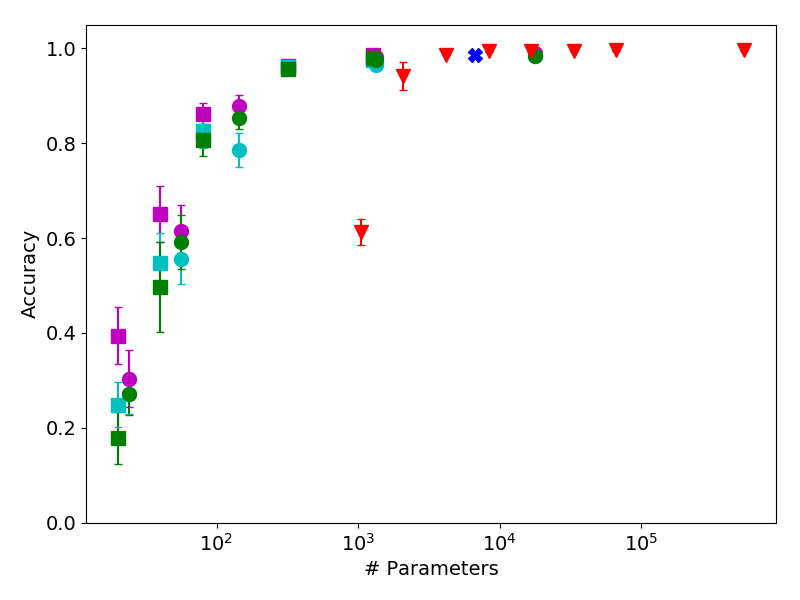}
\includegraphics[width=0.4\textwidth,keepaspectratio]{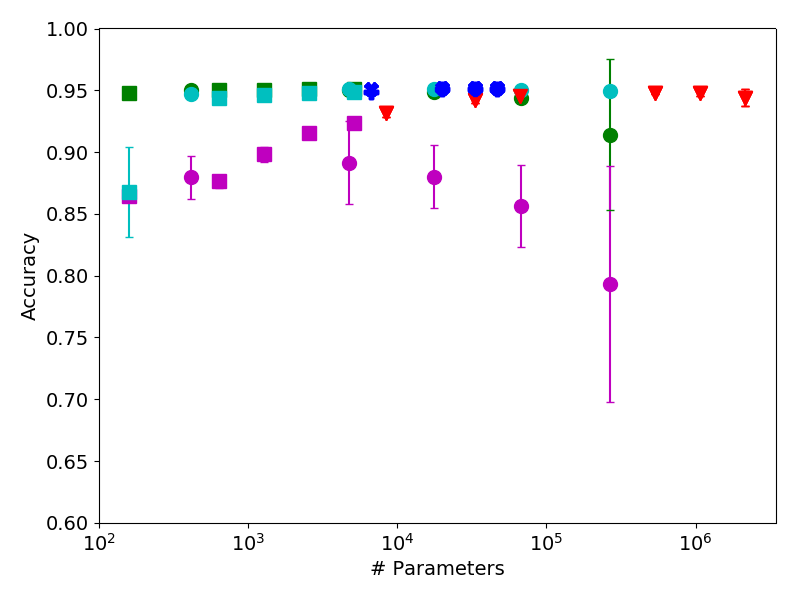}\\
\includegraphics[width=0.4\textwidth,keepaspectratio]{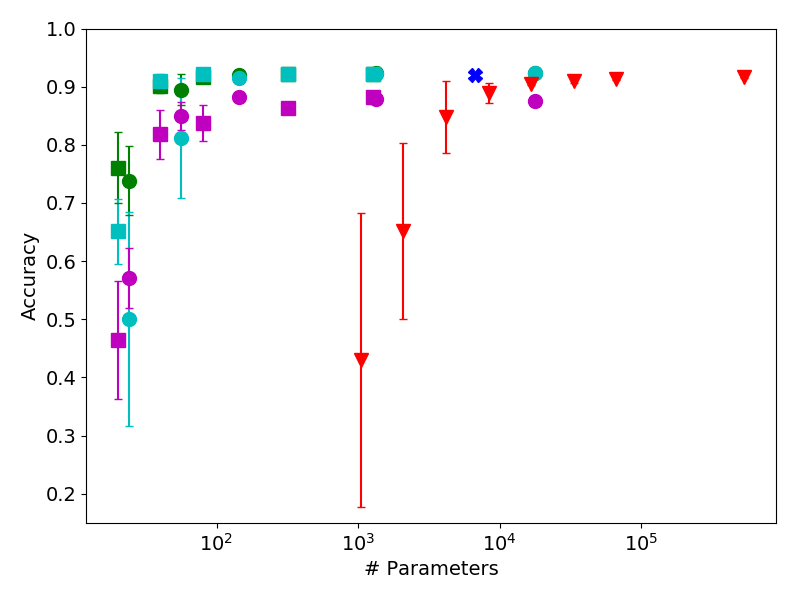}
\includegraphics[width=0.4\textwidth,keepaspectratio]{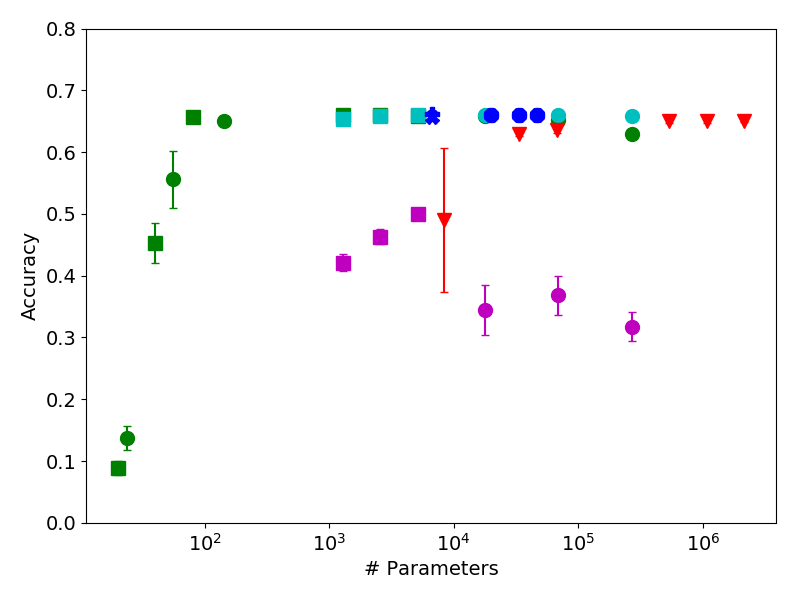}\\

\includegraphics[width=0.65\textwidth,keepaspectratio]{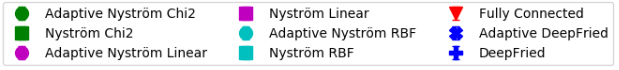} 
    \caption{Accuracy of models as a function of the number of parameters ($conv$ part not included) for various kernels on MNIST (top-left), SVHN (top-right), CIFAR10 (bottom-left) and CIFAR100 (bottom-right) datasets. RBF is a shorthand for Gaussian RBF.}
    \label{fig:resultsacc}
\end{figure*}

In a multiple kernels context, kernels $k^1,\ldots, k^l$ correspond to $l$  Nyström layers that can be computed in parallel then merged to encode the information provided by the different kernel representations. Learning the weights $\weightnys_1,\ldots,\weightnys_l$ in this case is, in a way, related to multiple kernel learning. This is a particularly interesting feature of our Nyström layer because it is sometimes difficult to know in advance which kernel will perform the best for a particular task, as demonstrated in Figure \ref{fig:resultsacc}. One possible merging strategy for the different kernel representations is simply to concatenate them such as:

\begin{equation}
\label{eq:mkl}
    \phi_{nys_{mkl}} = 
    \begin{bmatrix}
    \weightnys_1 \boldsymbol{k}^1_{\x, \nyssubsample} \\
    \vdots \\
    \weightnys_l \boldsymbol{k}^l_{\x, \nyssubsample}
    \end{bmatrix}
\end{equation}
with $\boldsymbol{k}^i_{\x, \nyssubsample}$ being the kernel vector obtained using the $i$\textsuperscript{th} kernel function $k^i$.

Comparing with \textit{Deep Fried Convnets} that also consists in replacing the top dense layers of a convolutional neural network by using kernel approximation, we mention two main structural differences with our model: 
(I) Nyström approximation has the flexibility to use any kernel functions and to combine multiple kernels while the \textit{Fastfood} approximation, used in the \textit{Deep Fried Convnets}, is limited to shift-invariant kernels, and (II) in contrast to \textit{Fastfood} the  Nyström approximation is data dependent. The storage ($O(d + dc)$) complexity of the \textit{Fastfood} approximation is lower than the Nyström approximation ($O(m^2 + mc)$) but we show in the experiments that this theoretical difference doesn't stand in practice, because the number of necessary examples is far lower than the dimension of the features in the output of the convolution blocks.

In the next section, we demonstrate the effectiveness of our method on some classical image classification datasets then we try to explore its possible extensions for few sample learning and multiple kernel learning, exploiting some well known advantages of kernel methods.




\begin{table*}[!h]
\centering
\begin{tabular}{|c|c|c|c|c|c|}
\hline
\textbf{Dataset} & \textbf{Input shape}        & \textbf{\# classes} & \textbf{Training set size} & \textbf{Validation set size} & \textbf{Test set size} \\ \hline
MNIST        & $(28 \times 28 \times 1)$ & 10                         & 40 000                       & 10 000                     & 10 000               \\ \hline
SVHN        & $(32 \times 32 \times 3)$ & 10                         & 63 257                       & 10 000                     & 26 032               \\ \hline
CIFAR10      & $(32 \times 32 \times 3)$ & 10                         & 50 000                       & 10 000                     & 10 000               \\ \hline
CIFAR100      & $(32 \times 32 \times 3)$ & 100                         & 50 000                       & 10 000                     & 10 000               \\ \hline
\end{tabular}
\caption{Datasets statistics}
\label{tab:data}
\end{table*}

\section{Experiments}
\label{sec:xp}

We present a series of experimental results that explore the potential of deep adaptive Nyström networks with respect to 
various classification settings.
First we consider a rather standard setting and compare our approach with standard models on image classification tasks. We explore in particular the behaviour of deep adaptive Nyström with various kernels and stress the very limited subsample size needed to achieve good accuracy performance. 
Next we investigate the use of Nyström layers in a small training set setting, which shows that our approach may allow to learn classes with only very few training samples, 
taking advantage of the reduced number of parameters learned by our model. Finally, we investigate the multiple kernel architecture and illustrate its interest when learning with RBF kernel to overcome the hyperparameter selection, and also we demonstrate the benefit of a multiple Nyström approach, combining kernels computed from individual feature maps.

Before describing all these results we detail the datasets used in our expriments.

\subsection{Experimental settings}


We conducted experiments on four well known image classification datasets: MNIST \cite{mnist}, SVHN \cite{svhn}, CIFAR10 and CIFAR100 \cite{cifar10}, details on these datasets are provided in Table \ref{tab:data}. We pretrained the convolutional layers using standard architectures on both datasets: Lenet \cite{lecun-98} for MNIST and VGG19 \cite{DBLP:journals/corr/SimonyanZ14a} for SVHN, CIFAR10 and CIFAR100. We slightly modified the filters' sizes in Lenet network to ensure that the dimension of data after the convolution blocks is a power of 2~(needed for the \textit{Deep Fried Convnets} architecture to which we compare). Those experiments focus on highlightling the potential of learning Adaptive Nyström and combination of them in a deep architecture. Learning jointly convolution layers and adaptive Nyström layer as an end-to-end strategy is also investigated and left for future work.





We compare three architectures in all conducted experiments. Pretrained convolutional parts are shared by the three architectures, which differ from the layers on top of it: (1) \textit{Dense} architectures use dense hidden layers, i.e. these are classical convnets architectures ; (2) \textit{Deep Fried} implements the \textit{Fastfood} approximation (Equation~\ref{eq:ff}) ;  (3) \textit{Nyström} stands for our proposal. 

For \textit{Dense} architectures, we considered one hidden layer with \textit{relu} activation function, and varied the output dimension as $\{2, 4, 8, 16, 32, 64, 128, 1024\}$ in order to highlight accuracies as a function of the number of parameters. Exploiting more hidden layers did not significantly increase performances, in our experiments. We hence let it to one in order to ease the readability of figures. For the \textit{Fastfood} approximation in \textit{Deep Fried Convnets} we consider that $\phiff$ is gained with  one stack of random features to form $\randcompff$ in equation~\ref{eq:ff}, except in the experiments of section~\ref{smallexpe} which yields a representation dimension up to 5 times larger. Regarding  our approach and $\phinys$, we varied the subset size $\nyssubsample \in \{2, 4, 8, 16, 32, 64, 128\}$, we compared linear, RBF and Chi2 kernels, and we chose as output dimension of $\weightnys$ the same size as $\textbf{K}_{11}^{-\frac{1}{2}}$, corresponding to the subset sample size. Finally we explored the adaptive as well as non-adaptive variants. 






Models were learned to optimize the cross entropy criterion with Adam optimizer and a gradient step fixed to $1e^{-4}$. 
By default the RBF bandwidth was set to the inverse of the mean distance between convolutional representations of pairs of training samples. All experiments were performed with Keras \cite{chollet2015keras} and Tensorflow \cite{tensorflow}.

The experiments below investigate the potential of our architecture. We do not aim to beat current state-of-the-art results on the datasets considered due to our limited resources with respect to the amount of necessary experiments. Consequently, we did not use tricks such as data augmentation and extensive tuning and, in particular, we did not use the best known convolutional architecture for each of the dataset, we rather used reasonable deep architectures: VGG19 \cite{DBLP:journals/corr/SimonyanZ14a} for the three datasets CIFAR10, CIFAR100 and SVHN and Lenet \cite{lecun1998gradient} for the MNIST dataset. 
We then fix a shared convolutional model (VGG19) and compare results gained with classification layers formed by either Nyström approximation, Fastfood approximation, or fully connected layers. 
Finally, although some work has been done on more advanced sampling schemes for the Nyström approximation, we adopted a stratified uniform sampling of the examples from the labeled training set as it has been shown in \cite{kumar2012sampling} to be a good cost/performance ratio sampling strategy.

\subsection{Exploring the potential of the method}

We compare now deep adaptive Nyström networks to  two similar architectures,  \textit{Deep Fried Convnets} and classical convolutional networks (inspired from VGG19 and Lenet depending on the dataset). We vary the number of parameters of each architecture in order to highlight classification accuracy with respect to needed memory space. 


Figure~\ref{fig:resultsacc} shows the compared networks accuracy with respect to the number of estimated parameters (e.g. variables) in the last representation layer plus the linear classification layer. Note that we ignore parameters for convolutions layers to ease the readability of figures since they are all equals in the compared models. We repeated each experiments 10 times and plotted average scores with standard deviations. 
Nyström results of increasing complexity (number of parameters) correspond to the use of a subsample of increasing size from 2 (leftmost point) to 128 (rightmost point). One may see that there is no need of a large subsample here. This may be explained since the convolutional part of the network has been learned to yield quite robust and stable representations of input images. We provide a figure in Section~\ref{sec:2D representations} to illustrate this (see Figure~\ref{fig:plot2d}).

Deep adaptive Nyström network is able to reach the same accuracy performance while using much fewer parameters than classical fully connected layers. Moreover, we also observe smaller variations that points out the robustness of our model. The flexibility in the choice of the kernel function is a clear advantage of our method, as illustrated, since the best kernel is clearly dependent on the dataset~(linear on MNIST, Chi2 on SVHN and CIFAR100, RBF on CIFAR10). 
We show for instance a gain by using the Chi2 Kernel~($k(\x_1, \x_2) = ||\x_1 - \x_2||^2 / (\x_1 + \x_2)$) that had been used for image classification  \cite{Vedaldi2012}. We also notice the benefit of adaptive variants of Nyström layer, suggesting that our model is able to learn a useful Kernel map, more expressive than one obtain from conventional Nyström approximation, and with fewer parameters than Fastfood approximation.

\begin{table*}[h!]
\centering
\footnotesize
\begin{tabular}{|l|c|c|c|c|c|c|c|c|}
\hline
 & \multicolumn{2}{|c|}{MNIST}   & \multicolumn{2}{|c|}{SVHN}   & \multicolumn{2}{|c|}{CIFAR10}  & \multicolumn{2}{|c|}{CIFAR100} \\
 & 5   & 20     & 5    & 20     & 5     & 20     & 5     & 20    \\
 \hline
Dense            & \textbf{49.7} (4) & 94.4 (0.5) & 65.6 (11.6) & 81.7 (3.9) & 39.1 (3.3) & 87.1 (3.7) & 19.2 (2.2) & 35.7 (2.7) \\
\hline
Adaptive-Deepfried   & 12.4 (3.3) & 12.4 (1.4) & 16.7 (5) & 21.0 (6.4) & 28.3 (9.2) & 41.2 (3.6) & 3.9 (1.2) & 6.4 (0.8) \\
\hline
Adaptive-Nyström-L  & 48.1 (5.5) & 95.0 (0.5) & 22.4 (6.9) & 29.6 (13.5) & 12.0 (5.6) & 27.8 (7.6) & 1.2 (0.6) & 1.9 (0.8) \\
\hline
Adaptive-Nyström-R  & 41.2 (7.7) & \textbf{95.5} (0.3) & 42.1 (29.6) & 53.5 (33.6) & \textbf{70.8} (4.4) & \textbf{92.2} (0.1) & \textbf{24.7} (2.6) & \textbf{62.1} (1.2) \\
\hline
Adaptive-Nyström-C & 26.4 (7.7) & 92.3 (1.8) & \textbf{89.6} (3.1) & \textbf{93.3} (1.3) & 67.1 (4.7) & \textbf{92.2} (1) & 20.2 (2.2) & 55.4 (1.9) \\
\hline
\end{tabular}
\caption{Classification accuracy of Dense layers architectures; Adaptive DeepFried, Deep Adaptive Nyström with linear (L), gaussian RBF (R), Chi2 (C) kernels, on small training sets with 5 and 20 training samples per class. Variance of results, computed on 30 runs, are given in brackets.}

\label{tab:SmallTrainingSet}
\end{table*}

\subsection{Learning with small training sets}
\label{smallexpe}
Here we explore the ability of our model to work with few training samples, from very few to tens of samples per class. It is an expected benefit of the method since the use of kernels could take advantage of small training samples. 

These preliminary experiments aim to show how the final layers of a convolutional model may be learned from very few samples, given a frozen convolutional model. We actually performed the following experiments by exploiting a trained convolution model that has been learned on the full training set and investigate the performance of deep adaptive Nyström as a function of the training set used to learn the classification layers. One perspective of this work is to exploit such a strategy for domain adaptation settings where the convolutional model is trained on a training set within a different domain as the classes to be recognized. 

Based on already trained convolution models, we leverage on the additional information  that one may easily include in our models, which is brought by the subsample set. Notice that this subsample may include unlabeled samples since their labels are not used for optimizing the model. 
Table~\ref{tab:SmallTrainingSet} reports the comparison of network architectures on four datasets. We consider \textit{Adaptive Nyström} using Linear, RBF or Chi2 kernels and compare with \textit{Dense} and \textit{Adaptive Deepfried} for training set sizes of 5, 10 and 20 samples per class. We only consider here adaptive variants since they brought better results than their non adaptive counterparts. 
We obtain models with different complexities: by increasing the hidden layer size in standard convolutional models, or by stacking the number of matrices $V$ in \textit{DeepFried} (up to 8 times, more was untractable on our machines), and by increasing the subset size for the Nyström approximation. Reported results are averaged over 30 runs.



\normalsize
One may see first that deep adaptive Nyström networks outperfom baselines on every setting except for 5 training samples per class on MNIST. The linear kernel performs well on MNIST but is significantly worse than baselines on harder datasets. At the opposite, deep adaptive Nyström with both RBF Gaussian kernel and Chi2 kernel significantly outperfom Adaptive DeepFried for all the datasets and perform equaly or significantly better than Dense architectures on the hardest CIFAR100 dataset. Intriguingly, the \textit{Deep Fried Convnets} performs particularly bad on this setting. Moreover one sees that no single kernel based Nyström representation dominate on all settings, showing the potential interest of combining multiple kernels as following experiments will show.

\begin{figure*}[h!]
\centering
\includegraphics[width=0.3\textwidth,keepaspectratio]{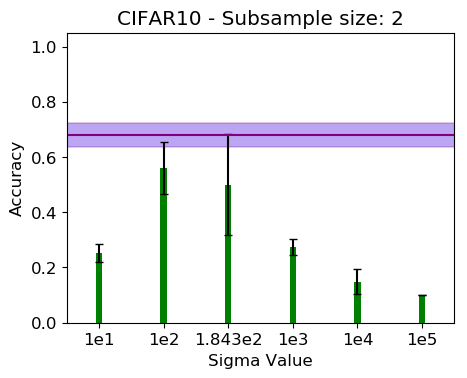}
\includegraphics[width=0.3\textwidth,keepaspectratio]{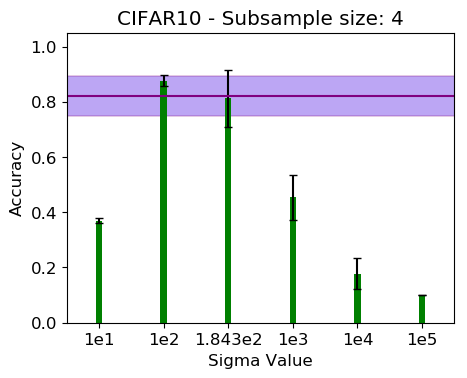}
\includegraphics[width=0.3\textwidth,keepaspectratio]{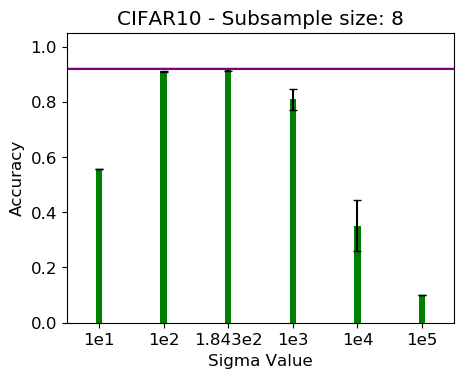}\\
\includegraphics[width=0.6\textwidth,keepaspectratio]{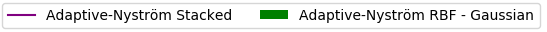} 

\caption{Comparison of \textit{Multiple Kernels}  that combines RBF kernels with various values of the bandwidth hyper-parameter. Multiple kernels performance is shown as an horizontal line while single kernel using one specific value of the bandwidth hyper-parameter $\sigma$. Plots correspond to a subsample size equal to 2 (left), 4 (middle) and 8 (right).}
        \label{fig:MKL}
\end{figure*}

\begin{table*}
\centering
\begin{tabular}{|l|c|c|}
\hline
\textbf{Model} & \textbf{Accuracy (std)}  & \textbf{Architecture} \\ 
\hline
Dense        & 68.0 (0.7) & 1 hidden layer 1024 neurons \\ 
\hline
Adaptive-Deepfried        & 67.6 (0.5) & 5 stacks   \\ 
\hline
Adaptive-Nyström       & \textbf{69.1} (0.2) & 256 subsamples + 512 Linear Kernels  \\ 
\hline
Adaptive-Nyström      & 67.6 (0.2) & 16 subsamples + 512 Chi2 Kernels  \\ 
\hline
\end{tabular}
\caption{\textit{Multiple Nyström} experiments on CIFAR100 obtained on top of VGG19 convolutions.}
\label{tab:multipledeepstrom}
\end{table*}

\subsection{Multiple kernel learning}
We report here results validating the multiple kernels strategies, that we described in section \ref{sec:deepstrom} and equation \ref{eq:mkl}. We conducted two different experiments:

First, 
we considered a combination of \gls{rbf} kernels with various bandwidths and for different subsample sizes.
In this case, each kernel function of $\{k^i\}_{i=1}^{l}$ is the \gls{rbf} function with a different sigma value. This multiple kernel strategy  exploits kernels defined with various values of the hyperparameter and allows to automatically handle tuning of hyper-parameter which usually requires heavy cross validation. Figure \ref{fig:MKL} shows the accuracy on  CIFAR10 dataset as a function of $\sigma$ value, where the performance of the multiple kernel Nyström is shown as a horizontal line. Plots report results for various subsample size equal to 2 (left), 4 (middle) and 8 (right), averaged over 10 runs. As may be seen, using our Multiple kernel network allows adapting the kernel combination optimally from the data without requiring any prior choice on the RBF bandwith.

Second, we investigated a variant of the Nyström architecture that we call \textit{Multiple  Nyström} approximations. Here we consider in parallel multiple Nyström approximations where kernels are dedicated to deal each with the output of a single feature map from the $conv$ layers. Let $conv(x)_i$ be the $i$\textsuperscript{th} feature map of $x$; in \textit{Multiple  Nyström}, we use $k^i(conv(x), conv(y)) = k(conv(x)_i, conv(y)_i)$.  Table~\ref{tab:multipledeepstrom} reports results on CIFAR100. We show the best performances obtained for each method by grid-searching on various hyper-parameters for each model, within a similar range of number of parameters. For \textit{Dense} model, we considered one or two hidden layers of 16, 64, 128, 1024, 2048 or 4096 neurons. \textit{Deepfried} is the adaptive variant where we varied the number of stacks in {1, 3, 5, 7}. We use deep adaptive Nystr\"om with subsamples of size {16, 64, 128, 256, 512}. We observe that both Nyström layers outperform the considered baselines, demonstrating the interest in combining Nyström approximations.

\subsection{2D representations}
\label{sec:2D representations}

\begin{figure*}[h!]
    \centering
\includegraphics[width=0.80\textwidth,keepaspectratio]{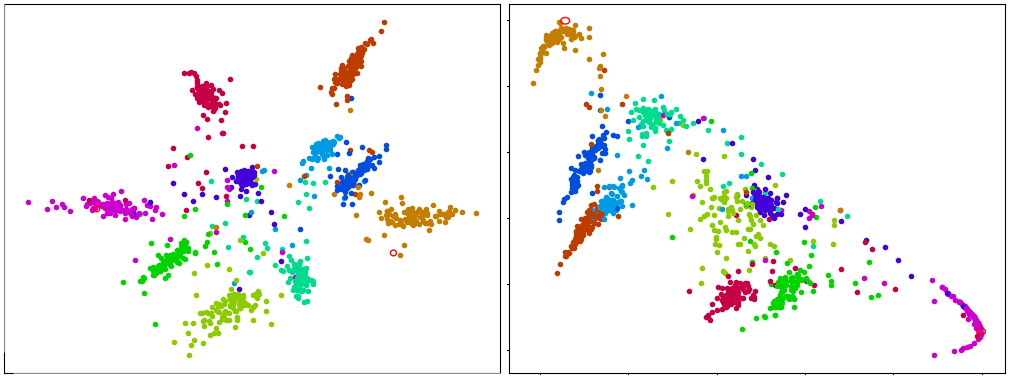}
    \caption{2-dimensional $\phinys$ representation of 1000 randomly selected test set samples from CIFAR-10 dataset, obtained with a subsample set of size 2 and a linear kernel (left) or Chi2 kernel (right). Each color represents a class.}
    \label{fig:plot2d}
\end{figure*}

Figure~\ref{fig:plot2d} plots the 2-dimensional $\phinys$ representations of  CIFAR10 test samples obtained with a subsample of size equal to 2 (while the number of classes is 10) and two different kernels. One may see here that the 10 classes are already significantly well separated in this low dimensional representation space, illustrating that a very small sized subsammple is already powerfull. Beside, we experienced that applying Nyström on lower level features output by lower level convolution blocks may yield good performance as well while requiring larger subsamples. 


\section{Conclusion}

We proposed deep Adaptive Nyström networks that can be used as a drop-in replacement for dense layers and hence form a new hybrid architecture that mixes deep networks and kernel methods. It is based on the Nystr\"{o}m approximation that allows considering any kind of kernel function in contrast to explicit feature map kernel approximations. 
 Our proposal reaches the accuracy performance of standard fully-connected layers while significantly reducing the number of parameters on various datasets, enabling in particular, learning from few samples. Moreover the method allows to easily deal with multiple kernels and with multiple Nyström variations.
 
\section*{Acknowledgement}

This work was funded in part by the French national research agency (grant number ANR16-CE23-0006). Finally, we would like to thank Ama Marina Kreme for her help and discussions.

\bibliographystyle{plain}
\bibliography{biblio}

 \newpage


\end{document}